\documentclass[runningheads]{llncs}
\usepackage{graphicx}
\usepackage{amssymb}
\usepackage{amsmath}

\usepackage{amsmath}
\usepackage{floatrow}
\usepackage{subfig}
\begin{document}

\title{Estimating the coverage in 3d reconstructions of the colon from colonoscopy videos}
\titlerunning{Estimating the coverage in 3d reconstructions of the colon}

\author{Emmanuelle Muhlethaler \thanks{Corresponding author} \orcidID{0000-0003-2766-6999} \and
Erez Posner \orcidID{0000-0003-2778-1612}\and
Moshe Bouhnik \orcidID{0000-0003-3003-2604}}

% index{Muhlethaler, Emmanuelle} 
% index{Posner, Erez} 
% index{Bouhnik, Moshe} 

\authorrunning{E. Muhlethaler et al.}
\institute{Intuitive Surgical, Inc. \\ 1020 Kifer Road, Sunnyvale, CA \\ \email{\{emmanuelle.muhlethaler, erez.posner, moshe.bouhnik\}@intusurg.com}}

% \institute{Intuitive Surgical, Inc., Sunnyvale, CA, USA \\ \email{\{firstname.lastname\}@intusurg.com}}
\maketitle

\begin{abstract}
Colonoscopy is the most common procedure for early detection and removal of polyps, a critical component of colorectal cancer prevention. Insufficient visual coverage of the colon surface during the procedure often results in missed polyps. To mitigate this issue, reconstructing the 3D surfaces of the colon in order to visualize the missing regions has been proposed. However, robustly estimating the local and global coverage from such a reconstruction has not been thoroughly investigated until now. In this work, we present a new method to estimate the coverage from a reconstructed colon pointcloud. Our method splits a reconstructed colon into segments and estimates the coverage of each segment by estimating the area of the missing surfaces. We achieve a mean absolute coverage error of 3-6\% on colon segments generated from synthetic colonoscopy data and real colonography CT scans. In addition, we show good qualitative results on colon segments reconstructed from real colonoscopy videos.

%\keywords{Colonoscopy \and Coverage \and Centerline.}
\end{abstract}

\section{Introduction}

Colorectal cancer is the third most common cancer worldwide \cite{cancerstats}. The early detection and removal of polyps during routine colonoscopy is an essential component of colorectal cancer prevention. The procedure is based on a visual examination of the colon using a single camera mounted on a flexible tube. During this procedure, 22\%-28\% of polyps are missed \cite{leufkens2012factors,polyp_miss_rate}, often because they never appeared in the field of view of the camera \cite{leufkens2012factors}. 

In recent years, efforts were made \cite{freedman2020detecting,rnnslam,reconstructionmodel,C3Fusion} to estimate the colonoscopy \textit{coverage}, that is, the fraction of the colon surface examined during a colonoscopy procedure. We define the coverage as the ratio: $\frac{S_{examined}}{S_{total}}$ where $S_{examined}$ is the area of the surface examined during the procedure and $S_{total}$ the area of the entire visible surface, including the missed regions. One possible approach to the coverage estimation problem is to compute a 3D reconstruction of the colon from the colonoscopy video \cite{rnnslam,reconstructionmodel,C3Fusion}. The missed regions will appear as holes in the reconstructed mesh. In this paper, we focus on the computation of the coverage given a 3d reconstruction of the colon. This has only briefly been addressed until now, as most works tend to focus on the 3d reconstruction itself. 

\begin{figure}
\includegraphics[width=\textwidth]{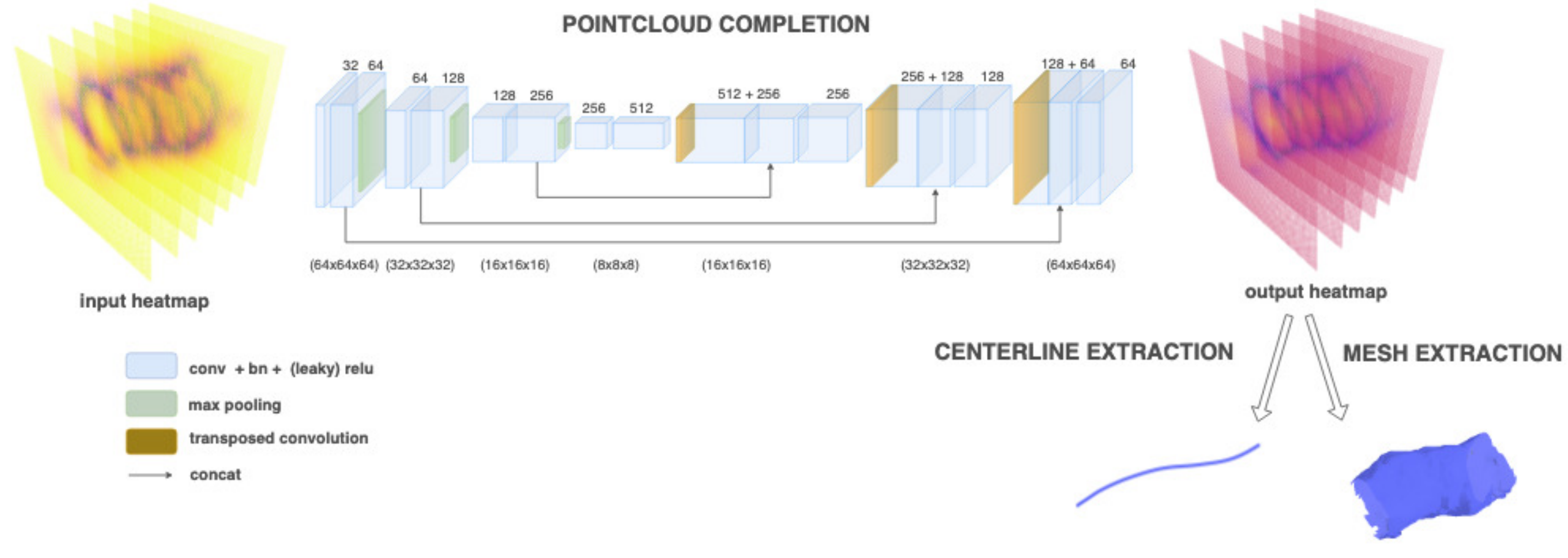}
\caption{The coverage estimation per segment, composed of three parts: Pointcloud completion, centerline extraction, mesh extraction.} \label{coverage_net_fig}
\end{figure}

We assume a reconstruction of the colon with holes and devise a method to estimate the coverage \textit{per segment}, where a segment is defined based  on the colon centerline. Estimating the coverage per segment provides a more detailed and useful information than a global colon coverage estimation. For instance, if the coverage estimation is run during a colonoscopy procedure, an estimation of the coverage per segment, rather than for the whole procedure, can allow the physician to easily identify the regions where the coverage is deficient and revisit the uninspected areas. We choose to base our method on the 3d completion of the reconstructed colon surface, thus providing an estimation of the location and shape of the missing surfaces, in addition to their area. Such an approach hasn't been explored yet, and makes our method easily interpretable, allowing us to visually assess the reliability of our coverage estimation when ground truth is unavailable. The central component of our method is a \textit{per segment coverage and centerline estimation module}, composed of 3 parts (see Fig.~\ref{coverage_net_fig}):  
\begin{enumerate}
    \item A \textbf{point completion network}, inspired by 3D-EPN \cite{dai2017complete}. It takes as input a heatmap representing a partial colon segment (i.e with holes) and outputs a heatmap of the completed segment together with its centerline.
    \item A \textbf{centerline extraction algorithm}, to extract the centerline from the estimated heatmap.
    \item A \textbf{mesh extraction algorithm}, to extract the surface mesh from the estimated heatmap.
\end{enumerate}

\section{Related works}

A number of works address the problem of constructing a 3d model of the colon from colonoscopy videos. In \cite{reconstruction2013}, the colon surface is generated based on folds detection and a depth from intensity model, but is limited to the reconstruction of single-frame segments. More recently, Ma et al. \cite{rnnslam} used a SLAM backbone based on DSO \cite{engel2016dso} together with a recurrent neural network for pose and depth estimation, to successfully reconstruct colon surfaces from real colonoscopy videos. They estimate the coverage on 12 real data segments, by mapping the surfaces onto a 2D rectangular frame, but no ground truth is available to measure the accuracy of the method. Zhang et al. \cite{reconstructionmodel} use a non rigid registration between a mesh model from a prior CT colonoscopy and a 3D reconstruction based on deep depth estimation and classic sparse features. This method is not applicable to most real life scenarios, where no CT scan is available. Posner et al.\cite{C3Fusion} use deep depth estimation and deep features to reconstruct 3D surfaces from colonoscopy videos. In contrast to other works, Freedman et al. \cite{freedman2020detecting} chose to avoid building a 3D reconstruction and instead train a number of networks to directly estimate the coverage from a sequence of images. This method provides an estimate of the coverage per segment, but has a few drawbacks, such as lack of interpretability and the fact that a segment is defined based on time (a fixed number of frames), and does not represent a physical colon segment of a given length.

We choose to base our method on the 3d completion of reconstructed colon pointclouds. The task of estimating complete 3D shapes from partial observations has many applications in computer vision and robotics. Recent solutions to the pointcloud completion problem can be roughly classified according to the type of deep architecture used. The earlier works are CNN based, and represent the pointcloud as a voxel grid \cite{dai2017complete,Han2017,Stutz_2018_CVPR}. An important limitation of this approach is the loss of resolution caused by the voxelization of the shape. Another approach consists of using a PointNet \cite{PointNet} type of architecture \cite{yuan2018pcn,topnet2019}, where a decoder reconstructs the complete pointcloud from a global learned feature. This process does not allow to clearly separate between the original points and the filled-up regions. In the completed shape, regions corresponding to the original pointcloud might have been distorted or lost details. To remedy this issue, \cite{skipatt2020} add a skip attention mechanism to the encoder decoder architecture.   

Our method also estimates the colon \textit{centerline}, as an intermediate step. The \textit{medial axis} or \textit{skeleton} of an object is the set of points having more than one closest point on the object boundary \cite{blum_67}. In the medical context it is also often called the \textit{centerline}, and in the case of a tubular object, it should consist of a single continuous line spanning the object. Some works \cite{auto,minpaths} address the issue of extracting the colon centerline from a CT scan of the colon with the purpose of generating an optimal trajectory for CT colonography . In \cite{minpaths} minimal paths are extracted from CT scans given 1 or 2 endpoints. 

\section{Coverage estimation of 3d colon reconstructions}

Our method estimates the coverage per segment of a 3D pointcloud representing a colon with holes. For maximum generality, we assume that our input consists only of a set of points $P = \lbrace \bold{p_{i}}\rbrace$, $\bold{p_{i}} \in\mathbb{R}^{3}$ with no further information.

We define a \textit{segment} using the centerline. The centerline is split into a number of continuous segments of a given arc length (e.g. $l=7 cm$). For each centerline segment, the corresponding colon segment is defined by two cross sections perpendicular to the centerline (see Fig.~\ref{colon_segmentation_fig}).

\subsection{Dataset}

\begin{figure}
\centering
\includegraphics[width=0.9\textwidth]{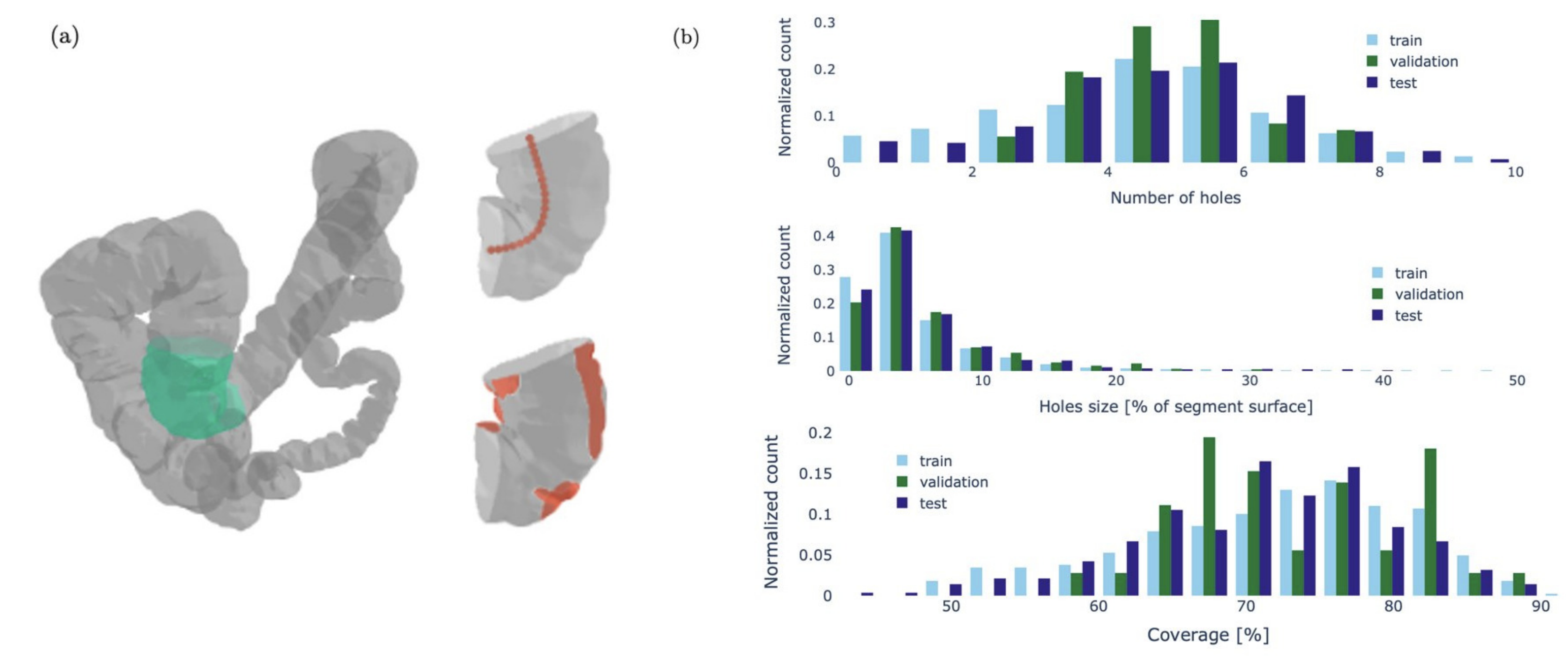}
\caption{(a) Colon from our CT scan dataset, together with a magnified cropped segment and its corresponding centerline and generated holes. (b) Distribution of holes number, sizes and coverage in our dataset. The individual holes sizes are expressed as a percentage of the segment surface.} \label{colon_segmentation_fig}
\end{figure}

Our training dataset is made of both CT and synthetic data. For the CT data, we used 3 colon meshes manually segmented from a dataset of colonography CT scans \cite{colonographydata}. As the resulting pointclouds do not contain any holes, we generated holes ourselves, by randomly cropping out spheres of various radii (see Fig.~\ref{colon_segmentation_fig} (a)). The generated distributions of coverage, holes numbers and holes sizes are displayed on the figure Fig.~\ref{colon_segmentation_fig}. The synthetic data consists of synthetic colonoscopy videos from which a reconstructed mesh can be generated. The various holes permutations were generated using a synthetic camera with random poses inside the colon. Using these meshes, we generated a dataset of colon segments to train and test our coverage estimation module. See Table~\ref{tab1}. For each segment in our train and validation set, 15 permutations corresponding to a different set of holes were generated. Each segment in our dataset also has a corresponding centerline. The centerlines were calculated on the full colon meshes using a classic skeleton extraction algorithm, in which the parameters were manually tuned for each mesh and the result was refined to obtain the desired properties (connectivity, centricity and singularity). These centerlines were used to split the original colons into segments (arc length $l\in [5, 6, 7]$) and to provide a GT centerline for our network training.

\begin{table}
\caption{Dataset summary. A, B represent the colons of our synthetic dataset and a, b, c the colons of our CT dataset.}\label{tab1}
\begin{tabular}{|c|c|c|c|}
\hline
& \textbf{train} & \textbf{validation} & \textbf{test}\\
%\hline
\hline
\begin{tabular}{@{}c@{}} \textbf{colon meshes} \\ {[a, b, c]} $\in$ CT scans; [A, B] $\in$ synthetic \end{tabular}
& a, b, A & a, b, A & c, B \\
\hline
\textbf{number of segments}  & 10200 & 3000 & 1200\\
\hline
\end{tabular}
\end{table}

While the overall shape of a colon varies enormously between people, we found that reducing our problem to segments (with scale, rotation, deformation and noise augmentations added during training) allowed for the training on one colon to generalize well to another, even when the overall shape differed greatly. We were able to generate a varied enough dataset with only a handful of individual colon mesh instances. As shown below, our method continued to work when our input was changed to a mesh reconstructed from real videos. 

The CT scans used in our dataset did not include significant irregularities in the colon shape, such as the ones caused by diverticula or extremely large polyps. The robustness of our method to such cases was not tested and our dataset might need to be augmented with these kind of irregularities in the future.

\subsection{Method}

\subsubsection{Pointcloud completion} Similarly to 3D-EPN \cite{dai2017complete}, we used a 3DCNN to complete a pointcloud represented by voxel grid. Although this type of approach suffers from a loss of resolution due to the voxelization of the shape, it is mitigated in our case by the following: (1) the full colon can consistently be split into small enough segments to get a satisfying resolution, (2) our main goal being coverage estimation, a loss of resolution is acceptable as long as it does not affect the coverage. We replaced the 3D-EPN architecture by 3dUNet \cite{3dUNet}, having observed that the fully connected layer of 3D-EPN \cite{dai2017complete} degraded the performance of our network. It might be related to the fact that, in contrast to classic shape completion networks, no object class needed to be encoded here. Our dataset consists of a single class of objects with a strongly constrained geometry. 

We defined a customized input and target representation for our problem, which is both easy to learn and from which a mesh and centerline with desired properties can be easily extracted. Our input is a 3D heatmap representing the partial colon segment (i.e. with holes). Our target is a 3D heatmap representing the centerline and completed segment. The heatmaps $\bold{H}_{input}$,  $\bold{H}_{target}$, are 64x64x64 voxel grids and are defined in the following way, for a voxel $\bold{v}$: 
\begin{eqnarray*}
\bold{H}_{input}[\bold{v}] &=& \tanh(0.2 * \mathrm{d}(\bold{v}, \bold{S_0}))\\
\bold{H}_{target}[\bold{v}] &=& \frac{\tanh(0.2 * \mathrm{d}(\bold{v}, \bold{S_1}))}{\tanh(0.2 * \mathrm{d}(\bold{v}, \bold{S_1})) + \tanh(0.2 * \mathrm{d}(\bold{v}, \bold{C}))}
\end{eqnarray*}

where $\mathrm{d}(\bold{v}, \bold{S_0})$ is the euclidean distance between $\bold{v}$ and the voxelized partial segment $\bold{S_0}$, $\mathrm{d}(\bold{v}, \bold{S_1})$ is the euclidean distance between $\bold{v}$ and the voxelized complete segment $\bold{S_1}$, and $\mathrm{d}(\bold{v}, \bold{C})$ is the euclidean distance between $\bold{v}$ and the voxelized centerline $\bold{C}$. The input heatmap is zero at the position of the (partial) segment surface and increases rapidly towards 1 away from it. The target heatmap is zero at the position of the (complete) segment surface, increases towards 1 at the position of the centerline, and converges to 0.5 everywhere else. Both heatmaps are illustrated in Fig.~\ref{coverage_net_fig}. We use an L2 loss to train the network.

\subsubsection{Centerline extraction}

% Incidently, we also noticed that including the centerline in our target heatmap improves the training when compared to using a heatmap of the complete mesh only.

\begin{figure}
\includegraphics[width=\textwidth]{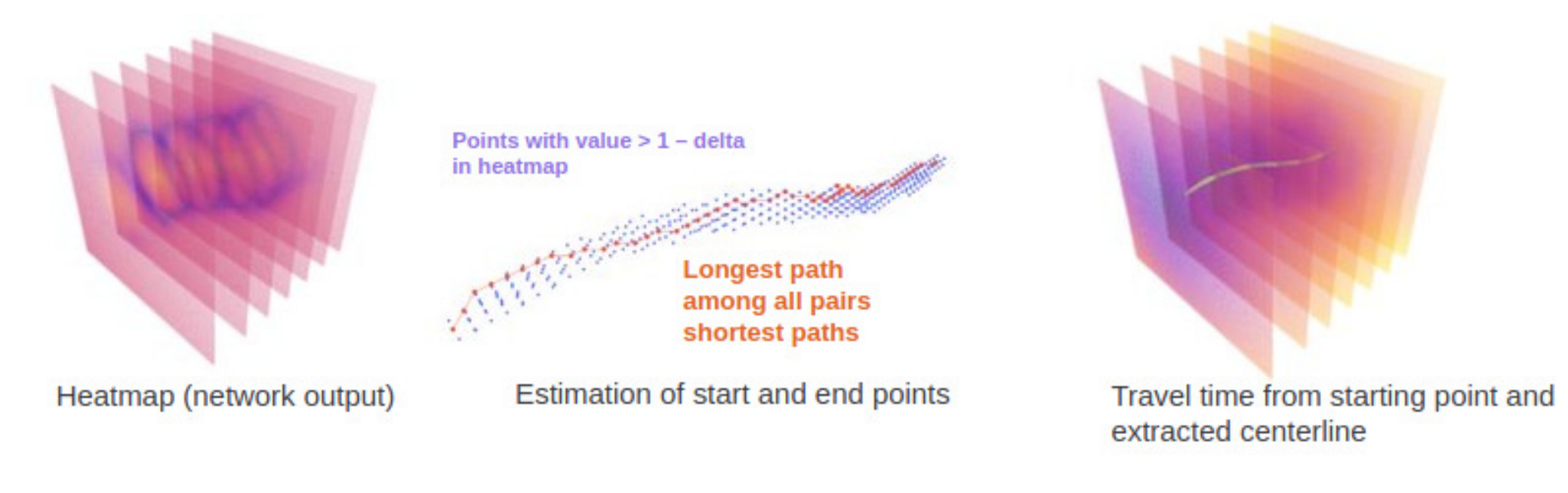}
\caption{Centerline extraction steps} \label{centerline_extraction_fig}
\end{figure}

The centerline is a key component of our pipeline. It allows us to split the colon into well defined segments and is also used for mesh extraction. Our output heatmap contains high values at the center of the colon, but simply thresholding the heatmap does not yield a singular and connected path. We use instead a minimal path extraction technique, similar to \cite{minpaths}. We add to it an initial step to calculate the start and end points of the centerline, which we don't know in general. To estimate the centerline start and end points, we create a nearest neighbor graph from the voxels with values $ > 1 - \delta$ in our heatmap (which roughly correlates to the centerline). The shortest path between all pairs is calculated and the longest path among them is selected. We use the extremities of this path as our centerline start and end points. We then compute the travel time from the starting point to each voxel in the volume, using the fast marching algorithm \cite{fastmarching} and our heatmap as speed map. The centerline is extracted by backpropagating the travel time from the end point down to the starting point. The different steps of our method are illustrated in Fig.~\ref{centerline_extraction_fig}.

\subsubsection{Coverage estimation}

Our estimation of the coverage includes 3 main steps: (1) Extract the completed pointcloud or mesh from the voxel grid, (2) differentiate between the filled up regions (the holes) and the rest of the pointcloud, (3) Calculate the ratio of the partial surface to the complete surface. 

The completed surface could be extracted by thresholding the predicted heatmap and extracting voxels with values close to 0. We found that with such a method, the extracted pointcloud can vary in thickness, making the calculation of the coverage (step 3) difficult. We opt instead to extract the completed mesh using marching cubes \cite{marchingcubes} in a neighborhood of the zeros-valued pixels. The surface to extract corresponds to a heatmap minima rather than an isovalue (with larger values on one side of the surface and smaller values on the other). We solve this issue by replacing the value of each voxel  $\bold{v}$ in the neighborhood of the surface by $\bold{v}_{new} = \mathrm{d}(\bold{s}, \bold{C}) - \mathrm{d}(\bold{v}, \bold{C})$, where $\bold{s}$ is the surface voxel closest to $\bold{v}$, $\mathrm{d}(\bold{s}, \bold{C})$ is euclidean distance between $\bold{s}$ and the centerline, and $\mathrm{d}(\bold{v}, \bold{C})$ is the euclidean distance between the $\bold{v}$ and the centerline. Once the completed mesh is extracted, holes are identified by comparison to the original partial pointcloud. We classify as belonging to a hole any vertex in the completed mesh with a distance to the partial pointcloud larger than $\frac{\sqrt{2}}{2}$ voxel size. The coverage is computed by dividing the partial mesh surface by the complete mesh surface. 

This per segment coverage estimation module can then be integrated into a broader pipeline, where the reconstructed colon is split into segments and the coverage is estimated per segment.

\section{Results}

% As shown below, our method is able to locally estimate the coverage as well as the location of holes with high accuracy. 

% We tested our coverage extraction module on 1200 segments from our test set as well as integrated to a synthetic pipeline where the local coverage is estimated during the reconstruction.

\begin{figure}
\includegraphics[width=0.46\textwidth]{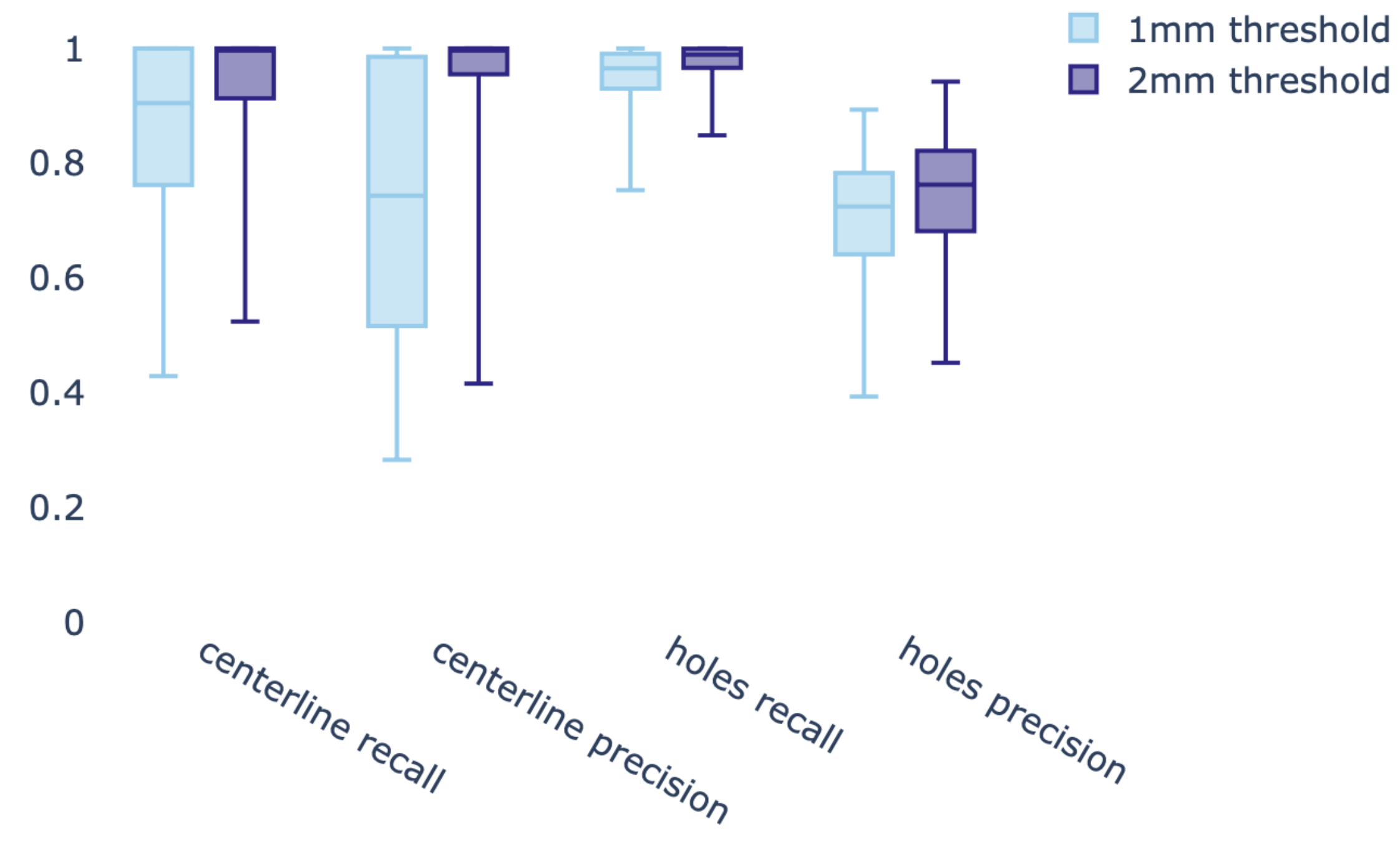}
\includegraphics[width=0.53\textwidth]{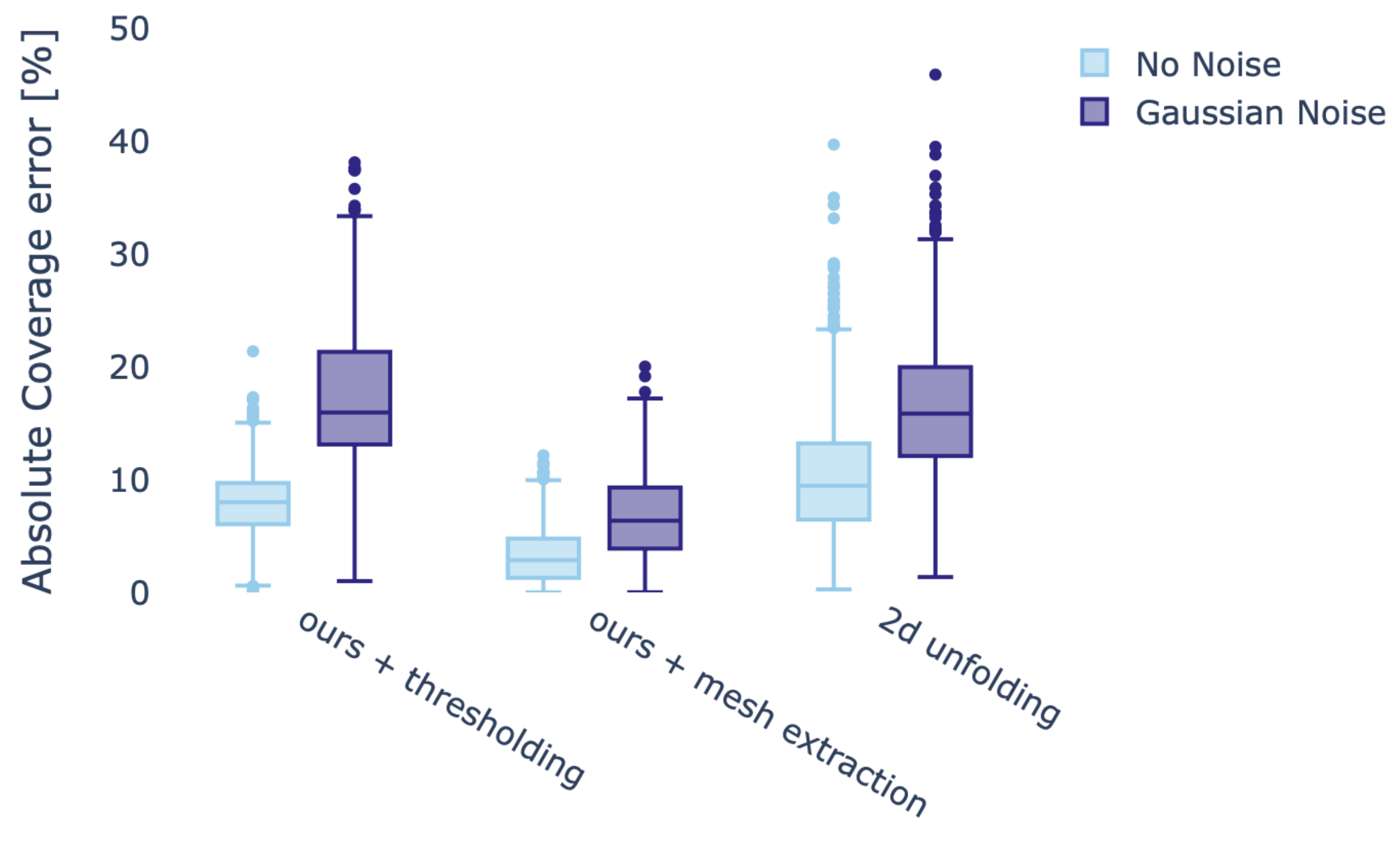}
\caption{Quantitative results on a benchmark of 1200 colon segments. Left: holes and centerline metrics. Lower whiskers correspond to the q5 quantiles. Right: Comparison of absolute coverage error for different methods and noises.}\label{coverage_error_plot}
\end{figure}

\subsubsection{CT and synthetic segments} 
\begin{figure}
\centering
\includegraphics[width=0.9\textwidth]{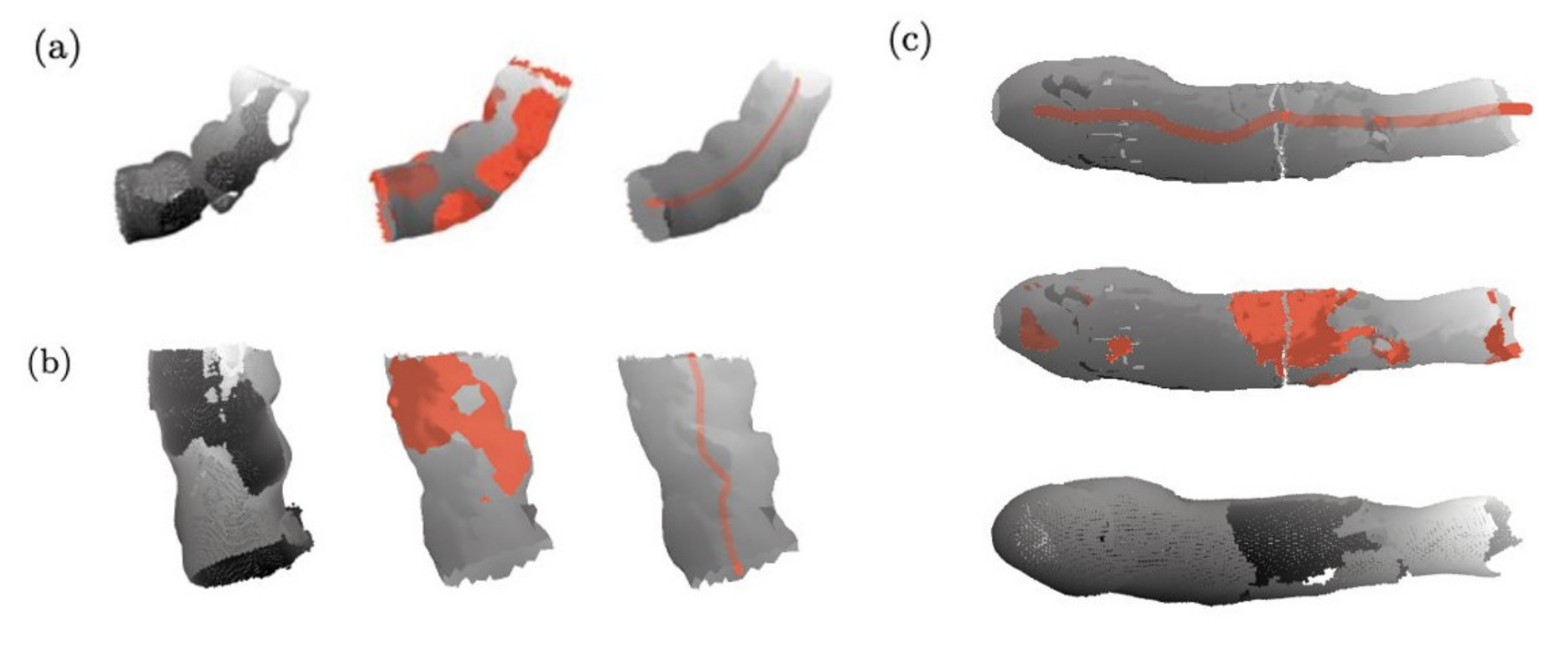}
\caption{Holes and centerline estimation on (a) CT scans segments, (b) 3d rigid colon print reconstruction, (c) colon 10K reconstruction. Each triplet represents: the input segment, the completed mesh with holes in red, the completed mesh with centerline in red. Additional examples are available in the supplementary material} \label{coverage_illustration_on_segments_fig:main} 
\end{figure}

We tested the coverage extraction module on 1200 segments from our test set. We used both regular noiseless data and segments to which gaussian noise ($\sigma = 0.5$mm) was added. We compared: (1) our coverage estimation method with the mesh extraction replaced by thresholding, (2) our full coverage estimation method and (3) the 2D unwrapping method described in \cite{rnnslam}. In \cite{rnnslam}, a straight line was used as centerline, which is not possible in general on curved segments. We chose to use our learned centerline instead. As shown on Fig.~\ref{coverage_error_plot}, we obtain the lowest absolute coverage error (3\% and 6\% MAE respectively on noiseless and noise augmented data) when using our learned heatmap together with mesh extraction. We observed that all methods tended to be biased towards underestimating the coverage, i.e. overestimating the surface of the holes. In the case of ours + mesh extraction method, it seems that the main reason for this bias is the detection of nonexistent holes at the extremities of the segment (see Fig.~\ref{coverage_illustration_on_segments_fig:main}). These errors are usually removed when we have access to the surface of adjacent segments.

We evaluate our centerline and holes estimation using the metrics: precision = $\mathrm{mean}_{\bold{p}_{est}}(\mathrm{min}_{\bold{p}_{gt}}||\bold{p}_{est} - \bold{p}_{gt}|| < th)$, recall = $\mathrm{mean}_{\bold{p}_{gt}}(\mathrm{min}_{\bold{p}_{est}}||\bold{p}_{est} - \bold{p}_{gt}|| < th)$, where $th \in [1 \mathrm{min}, 2 \mathrm{mm}]$ and $\bold{p}_{gt}$, $\bold{p}_{est}$ are respectively GT and estimated 3d points. The choice of thresholds is both related to the resolution of our heatmap (around 1mm for a 6cm segment voxelized into a 64x64x64 voxel grid) and to the fact that 1mm corresponds to the lowest range of polyp sizes \cite{polyp_size}. We achieve very high precision and recall, in particular for the centerline, with a median value of 1.0 for both precision and recall. We observed that most of our outliers could be traced back to a wrong estimation of the centerline extremities. Training our network with a stronger emphasis on the centerline might mitigate this issue.

\subsubsection{Rigid colon print and real data}\label{real data} We additionally tested our method on surfaces reconstructed from (1) a video recorded using a colonoscope of a colon 3d rigid print \cite{C3Fusion},  (2) a real optical colonoscopy video from the Colon10K dataset \cite{colon10k}. In both cases, the reconstruction was obtained using the method described in \cite{C3Fusion}. We obtained good qualitative results (see Fig.~\ref{coverage_illustration_on_segments_fig:main}). The segments obtained by running the reconstruction \cite{C3Fusion} on the real colonoscopy sequences \cite{colon10k} didn't exhibit any holes. This is due to both the data and the reconstruction algorithm. On the data side, the colon segments of \cite{colon10k} tended to be particularly smooth, with very small haustral folds. On the reconstruction side, the deep monocular depth estimation tended to smooth out discontinuities, a known phenomena \cite{monodepth2}. To obtain reconstructions with enough holes to test our method, we used 3 different subsets of the original frame sequences. We generated this way 3 sets of mesh with holes. Using the complete sequence reconstruction as ground truth for the full mesh (without holes), we obtained the following absolute coverage errors of: 5.2\%, 3.8\% and 7.6\%.  We found our method slightly more prone to errors when applied to this data, especially in cases where the reconstruction is noisy or contains errors. Training our network using more realistic noise augmentations and/or some actual reconstruction data (e.g. reconstruction from synthetic data colonoscopy) might help making our method more robust to these kind of failures.

\section{Conclusion}

We presented a novel method for estimating coverage given a 3d reconstruction of a colonoscopy procedure. Our method can be used to provide robust and interpretable local coverage feedback during a colonoscopy procedure, with 3D visualization of the missed surfaces.

\bibliographystyle{splncs04}
\bibliography{bibliography}

\newpage

\section{Supplementary Material}

\begin{figure}
\includegraphics[width=0.5\textwidth]{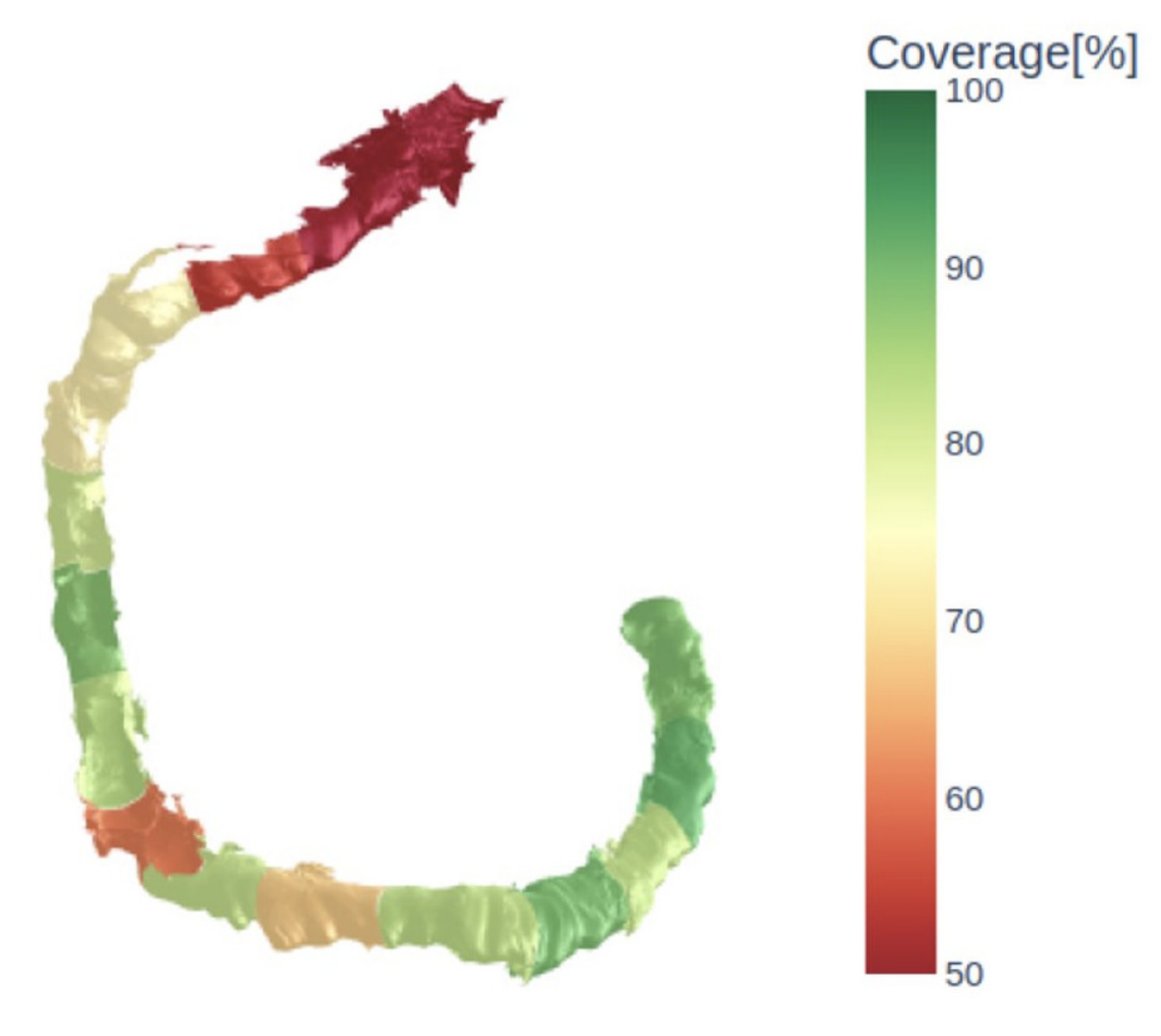}
\caption{Illustration of the colon 3d rigid print reconstruction and coverage estimation. Segments are colored according to their coverage, on a color scale ranging from red to green, where 50\% coverage or less corresponds to red.} 
\end{figure}

\begin{figure}
\includegraphics[width=0.5\textwidth]{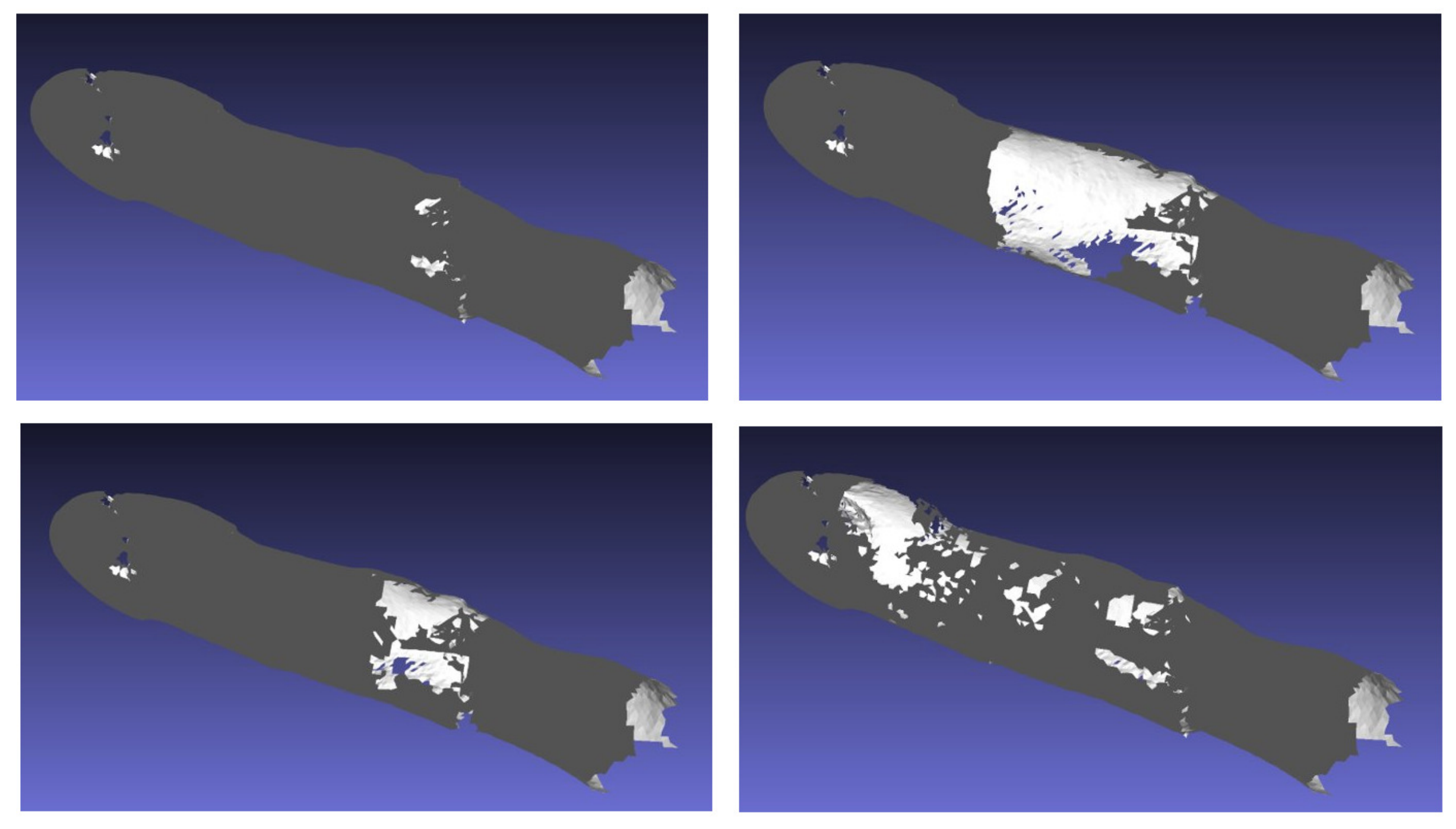}
\caption{Our colon 10K data. The upper left image is the full sequence reconstruction. The others are partial reconstruction, using only a subset of the sequence frames.} 
\end{figure}

% Work-around for bug in caption3.sty v2.0..2.2d regarding \sidesubfloat
\makeatletter
\let\caption@@@make@ORI\caption@@@make
\def\caption@@@make{%
  \caption@ifundefined\caption@lfmt{\let\caption@lfmt\caption@labelformat}{}%
  \caption@ifundefined\caption@fmt\caption@format\relax
\caption@@@make@ORI}
\makeatother

\begin{figure}
\centering
\sidesubfloat[]{\label{main:A}\includegraphics[width=0.9\textwidth]{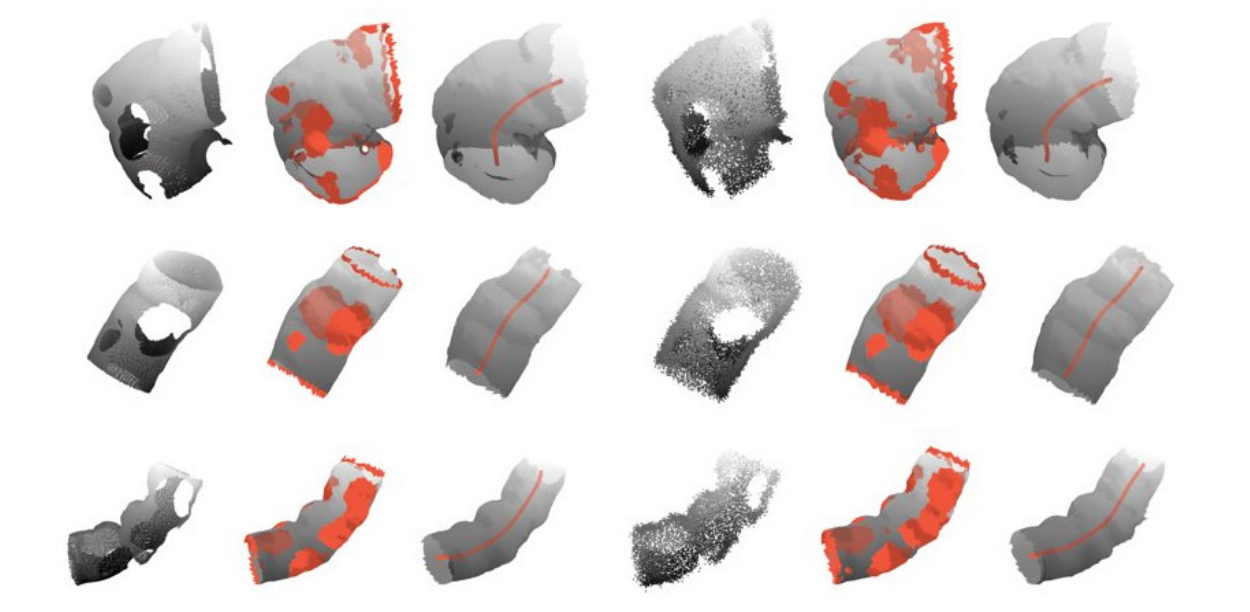}}\\
\bigskip
\sidesubfloat[]{\label{main:B}\includegraphics[width=0.9\textwidth]{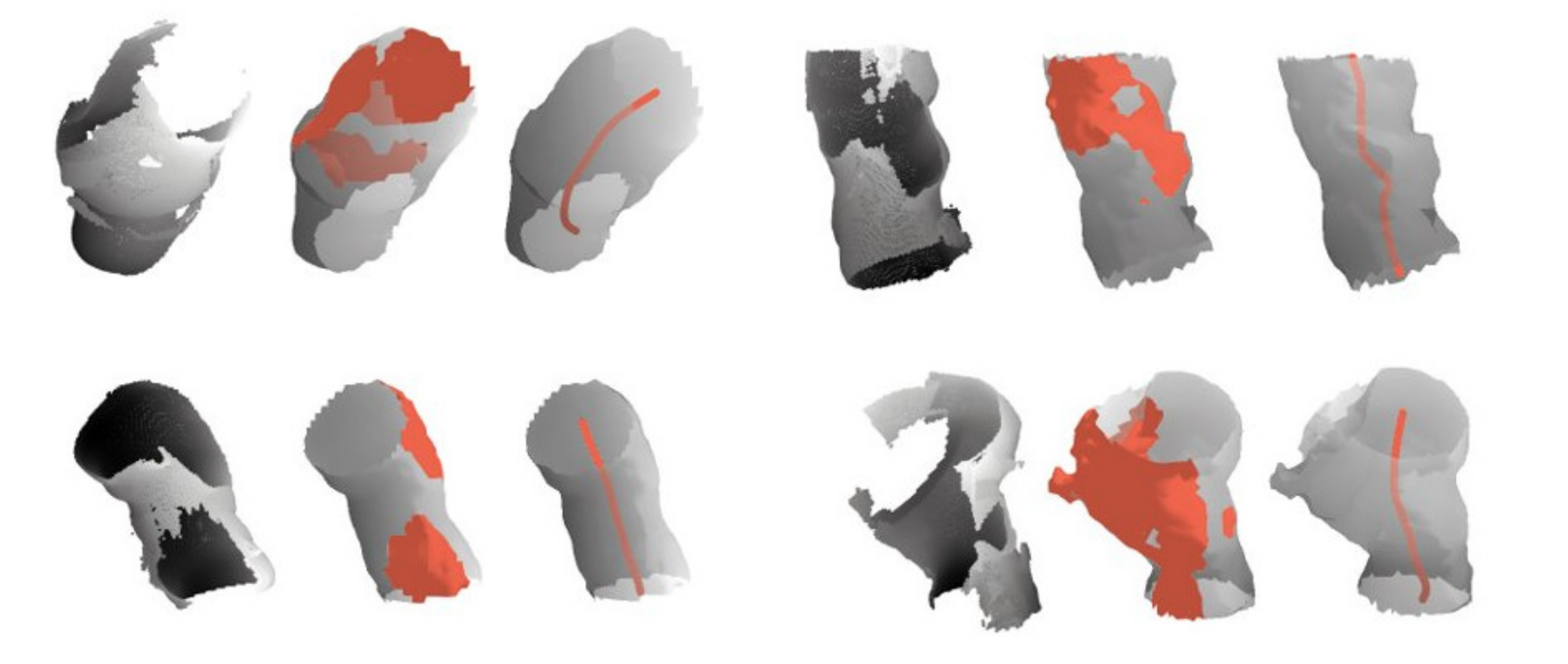}}\\
\bigskip
\sidesubfloat[]{\label{main:C}\includegraphics[width=0.9\textwidth]{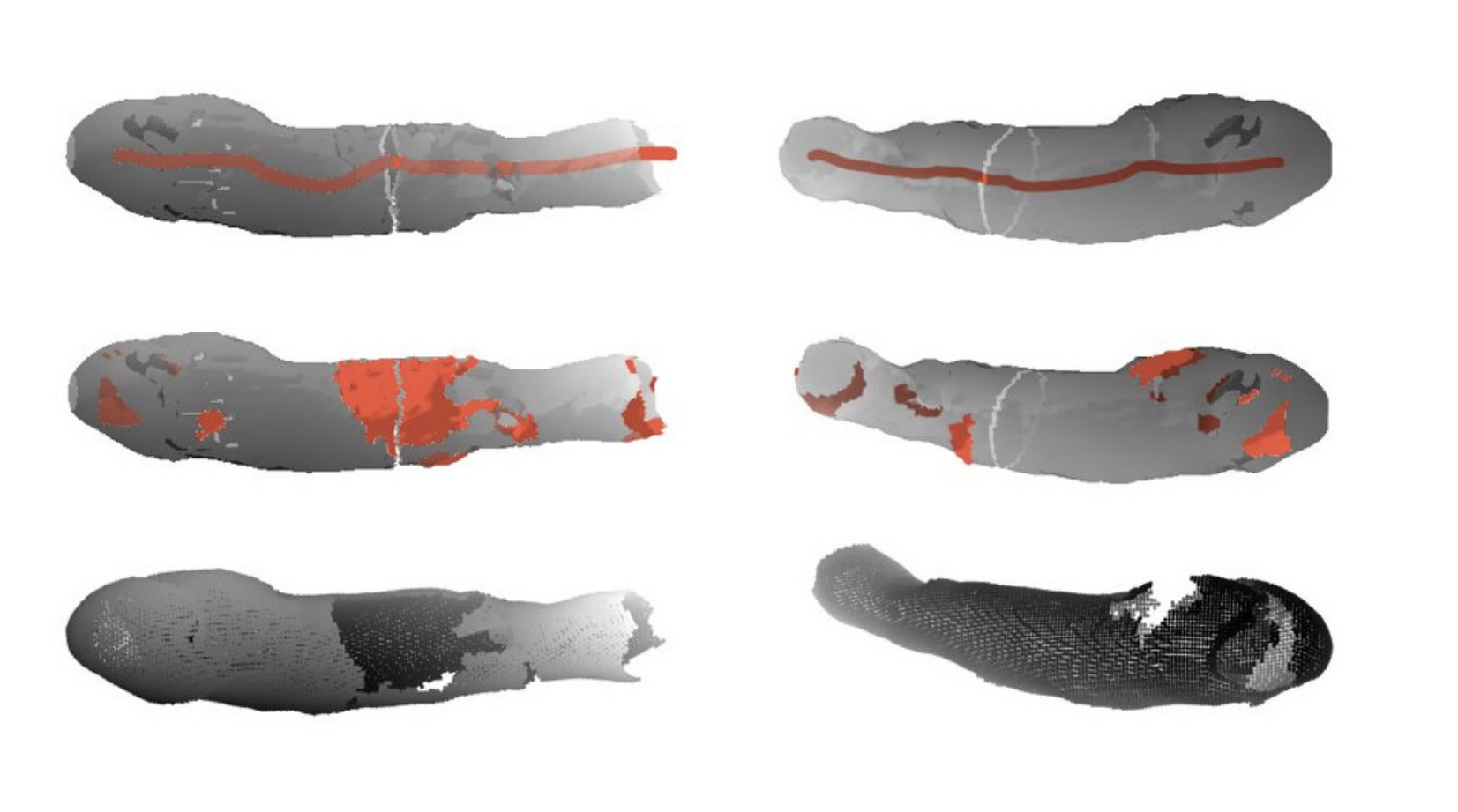}}
\caption{Triplets illustrating our holes and centerline estimation on different datasets. Each triplet represents: the input segment, the completed mesh with holes in red, the completed mesh with centerline in red. (a) Segments from the synthetic data test set. On the left, the input is noiseless, on the right, noise augmentations were added. (b) Segments from colon rigid print reconstruction. (c) Segments from real colonoscopy (colon 10K) video reconstruction.} 
\end{figure}

\end{document}